%% file: Template.tex
\documentclass{article}
\usepackage{spconf,amsmath,graphicx}

\input{math_commands}

\title{ Improved Projection Learning for Lower Dimensional Feature Maps}

\twoauthors
 {Ilan Price}
	{Mathematical Institute\\
	University of Oxford\\
	\& The Alan Turing Institute}
 {Jared Tanner}
	{Mathematical Institute\\
	University of Oxford\\}
\begin{document}
%
\maketitle
\begin{abstract}
The requirement to repeatedly move large feature maps off- and on-chip during inference with convolutional neural networks (CNNs) imposes high costs in terms of both energy and time. In this work we explore an improved method for compressing all feature maps of pre-trained CNNs to below a specified limit.  This is done by means of learned projections trained via end-to-end finetuning, which can then be folded and fused into the pre-trained network. We also introduce a new `ceiling compression' framework in which evaluate such techniques in view of the future goal of performing inference fully on-chip.
\end{abstract}
\begin{keywords}
efficient deep learning, convolutional neural networks, feature map compression
\end{keywords}

\section{Introduction}

Modern neural network architectures can achieve high accuracy while possessing far fewer trainable parameters than had traditionally been expected.  Prototypical examples include  compact weights and shared parameters in convolutional and recurrent neural networks respectively, as well as sparsifying and quantizing the weights within such networks, see \cite{blalock2020state,quantization_survey2022} and references therein.  The efficiencies in storing and transmitting these networks stand in stark contrast to their efficiency at inference time which is determined not only by the size of the model itself (weights, biases, etc), but increasingly importantly by the intermediate feature-maps (representations) generated as the outputs of successive layers and inputs to the following layers.  For example, with imagenet resolution (224x224) inputs, and even without sparsifying the networks, the single largest feature map is 7\% of the model size in Resnet18, 2.3\% in VGG16, and 34\% in MobilenetV2.  The relative model to feature map sizes can be dramatically exacerbated as these networks can have their number of parameters reduced to only a tiny fraction of their original size without loss in classification accuracy \cite{blalock2020state}. This motivates a line of research to improve efficiency by compressing feature maps too; see the related works Section~\ref{sec:related work}.

Typically, however, not all feature maps need to be stored simultaneously - once a feature map has been used as the input for the following layer(s), it can be immediately deleted.  Herein we propose an improved method for learning low-rank projections which can be incorporated into pre-trained CNNs to reduce their maximal memory requirements.  So doing, this approach seeks to both reduce the memory requirements on a device, and ideally to eliminate off-chip memory access mid-forward-pass, which can dominate power usage \cite{sze2017efficient, yang2017method}, a goal which is would enable lower-power, edge-device deployed deep networks. 

\begin{figure*}[h!]
    \centering
    \includegraphics[width=\textwidth]{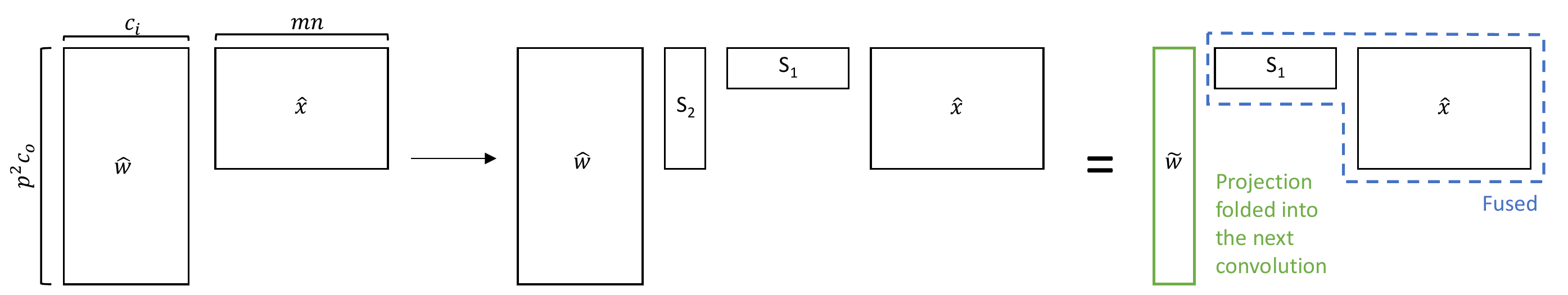}
    \caption{Illustration of the learned projections $S_1$ and $S_2$, which correspond to a projection of both the convolutional kernel reshaped as $\hat{w}$ and the post-activation feature map reshaped as $\hat{x}$.  The projection matrix $S_2$ can be folded into weight matrix $\hat{w}$ to form a lower dimensional weight matrix $\tilde{w}$.}
    \label{fig:matrices figure}
\end{figure*}

\section{Related work}\label{sec:related work}
One strand of research on feature map compression makes use of (and often tries to increase) the sparsity of post-activation feature maps, which endows a natural compressibility. Works such as  \cite{parashar2017scnn}, \cite{han2016eie}, and \cite{aimar2018nullhop, rhu2018compressing}, develop accelerators which use Zero Run-length encoding, compressed column storage, and zero-value compression, respectively, to leverage the naturally occurring feature map sparsity to shrink memory access. 

In \cite{cavigelli2019ebpc, rutishauser2020fly}, the authors leverage both `sparsity' and `smoothness' of feature maps, by decomposing the streamed input into zero and non-zero stream, and applying run-length encoding to compress the the lengthy runs of zeros. They then compress the non-zero values with bit-plane compression.  Similarly, \cite{georgiadis2019accelerating} induces sparse feature maps by adding $L_1$ regularisation when finetuning pre-trained networks. Furthermore, they use linear (uniform) quantisation of the feature maps,  as well as entropy coding for the resulting sparse feature maps. Lastly, \cite{kurtz2020inducing} proposes an alternative method for inducing sparse activations, based on finetuning with Hoyer-sparsity regularisation and a new discontinuous Forced Activation Threshold ReLU (FATReLU) defined for some threshold $T$ as $0$ if $x<T$ and $x$ otherwise.  Given that the sparsity of the feature maps of the aforementioned methods is input dependent, the compressibility and resulting compute resources needed for inference with this class of methods are not guaranteed or known ahead of time.

An alternative line of work focuses on transform-based compression. In \cite{shi2021transform}, they propose applying a 1D Discrete Cosine Transform (DCT) on the channel dimension of all feature maps, which are then masked and then zero value coded. Similary, \cite{shao2021memory} uses a combination of a 2D DCT transform, sparsification, and quantisation by removing high-frequency components, and sparse matrix encoding. Other works have proposed using PCA on a sample batch to pre-compute the transform \cite{chmiel2020feature}. The basic idea is to flatten the feature maps into a matrix and then orthogonalise the channel vectors. In \cite{xiong2020stc} only the top $k$ components are kept, ($k$ being determined by a set threshold for what percentage of $\sum_i \sigma_i^2$ must be captured).

Most similar to the present work is the Compression Aware Projection (CAP) method proposed in \cite{tai2022compression}, which, like ours, inserts learnable projections into trained networks. There are, however, a number of notable differences between that work and what is presented here, including different loss functions for training the projections, different means for selecting the projection dimensions for each feature map as well as, when and where to insert the projections. That work also chooses to compress pre-activation rather than post-activation feature maps, which -- as discussed in Section~\ref{sec:experiments} -- has a substantial impact on the effective compression factor.

\section{Learning feature map and kernel projections}\label{sec:method}

Denote a single feature map as $x \in \R^{c\times m \times n}$ where $c,m,n$ are the number of channels, width, and height respectively. Let $x^{(i)}$ denote the output of the $i$th layer, and input to the $(i+1)$th layer. We denote convolutional weight tensors $w \in \R^{c_o \times c_i \times p \times p}$, where $c_o, c_i, p$ denote the number of output channels (i.e. number of filters), number of input channels, and the spatial size of each filter, respectively. We denote convolution with the $*$ operator. We can consider feature maps to be either pre- or post- activation function $\phi$, that is, we may have $x^{(i)} = w^{(i)}*\phi(x^{(i-1)})$ or $x^{(i)} = \phi(w^{(i)}*x^{(i-1)})$. Unless otherwise specified we consider post-activation feature maps, as this yielded significantly better overall compression in our preliminary experiments.

Our goal is to compress the feature maps $x^{(i)}$ to a compressed representation $y^{(i)}$, in such a way that $y^{(i)}$ can be computed directly without ever storing the full $x^{(i)}$. We propose a linear auto-encoder based model, applied to a reshaped feature map $\hat{x} \in \R ^{d_1 \times d_2}$. To be precise, we will learn a linear projection matrix $S_1 \in \R^{k \times d_1}$ and lift matrix $S_2\in\R^{d_1 \times k}$, $k<d_1, d_2$, such the the compressed representation is $y^{(i)} = S_1\hat{x}^{(i)}$, which can then be decompressed in a streamed fashion by appropriately computing $S_2 y^{(i)}$. These learned projections will be inserted into a pre-trained model, whose entries are held fixed, with only the learned projection trained, and then eventually fused into the existing layers where possible for efficiency. Experimentally we achieve best results when projecting the channel dimension of $x^{(i)}$, and thus $d_1=c$ and $d_2= m \times n$. The resulting layer computation is equivalent to 
\begin{align}\label{eq: post-activation sketch layer}
    x^{(i)} = \phi(w^{(i)} * R'(S_2^{(i-1)} S_1^{(i-1)} R(x^{(i-1)}))),
\end{align}
where $\phi$ is the activation function, and $R$ and $R'$ denote reshaping the feature map tensor to a matrix and back again, respectively. Fortunately, as noted in Section \ref{subsec: no overhead}, these reshapes can be avoided in practice by fusing the projection with the previous layer, and folding the lift in to the next convolution as $\tilde{w}^{(i)}:=w^{(i)} * R'(S_2^{(i-1)})$. 

\subsection{Interpretations and choice of training objective}
The matrix $S_2S_1$ is low rank, and thus we can interpret the impact of inserting it after an activation function as replacing the next layer's weights $w$ with a low rank approximation. To see this, consider reshaping $w$ to $\hat{w} \in \R^{p^2c_o \times c_i}$, in which case the convolution $w * x$ can be cast as a simple function of the matrix product $\hat{w}\hat{x}$, which becomes $\hat{w}S_2S_1\hat{x}$ after inserting the projections, as illustrated in Figure \ref{fig:matrices figure}. Optimal low rank approximation is often defined in terms of the L2 reconstruction error. Under this metric we have analytically optimal $S_1$ and $S_2$. Specifically, to minimise the reconstruction error of the weights
\begin{align}
    \argmin_{S_1, S_2} \| \hat{w}S_1S_2 - \hat{w}\|_2 = U_k\Sigma_kV_k^\top
\end{align}
where $U_k\Sigma_kV_k^\top$ denotes the truncated SVD of $\hat{w}$, taking $S_1 = S_2^\top = V_k$ is optimal, since, $\hat{w}S_1S_2 = U\Sigma V^\top V_k V_k^\top = U_k\Sigma_kV_k^\top$. 

Alternatively, we could view our approach as multiplying each feature map $\hat{x}$ with a low rank matrix from the left. Minimising the average reconstruction error of the resulting (low rank) feature maps would aim to solve
\begin{align}
    \min_{S_1, S_2} \| S_1S_2\hat{x}^{(i)} - \hat{x}^{(i)}\|_2,
\end{align}
in expectation over the dataset. This explains approaches like the PCA based approaches \cite{chmiel2020feature} as well as the hint-loss \cite{zhang2015efficient} used in \cite{tai2022compression}.  Moreover, \cite{tai2022compression} also propose the network's output vector \textbf{\textit{o}} should remain unchanged with the compressed-feature-maps network, and so suggest an additional loss term $KL($\textbf{\textit{o}} $\|$ \textbf{\textit{o$'$}}$)$, where \textbf{\textit{o$'$}} is the compressed-network's output.

However, the true goal is not optimal approximation of either the weights or the feature maps, or even the network outputs. Rather, we aim for reconstructions which results in optimal accuracy on the original task, i.e. the test error. Thus instead of taking the analytically optimal weights projection, or training a standard auto-encoder on each layers feature maps, we learn $S_1$ and $S_2$ by finetuning the network using the original training data and loss functions, but with all other model parameters held fixed. This will allow us to jointly learn the subspaces into which both the weights and feature maps can be projected before computing the convolution which incur minimal decrease in accuracy.

\subsection{No storage overhead from projection matrices}\label{subsec: no overhead}

As the goal of this method is to shrink the storage requirements for inference, we need to account for any overhead due to storing the projection matrices in each layer. Fortunately, the lift applied by $S^{(i)}_2$ can be cast as a $1\times1$ convolution, which can be folded into the next convolution of $w^{(i)}$ as $\tilde{w}^{(i)}:=w^{(i)} * R'(S_2^{(i-1)})$, resulting in a single convolution being applied to the compressed input, with a new kernel which is smaller than $\hat{w}^{(i)}$ (see Figure \ref{fig:matrices figure}). So long as $k<\frac{p^2c_oc_i}{p^2c_0 + c_i}$ -- as is the case in all of our experiments -- the number of elements in $\tilde{w}^{(i)}$ plus the number of elements in $S_1^{(i)}$ will be less than the number of elements in $w^{(i)}$. 

\section{Experiments}\label{sec:experiments}

Here we conduct a study focusing solely on introducing the proposed learned projection as shown in Figure \ref{fig:matrices figure} (absent any quantisation and encoding), so as to best understand the possible levels of compression attributable thereto. The addition of quantisation and encoding schemes will yield further compression gains and should be investigated in future work.  Our experiments focus on Imagenet image classification, which is the standard benchmark task in prior related work. 

\begin{figure}[h!]
    \centering
    \includegraphics[width=0.5\textwidth]{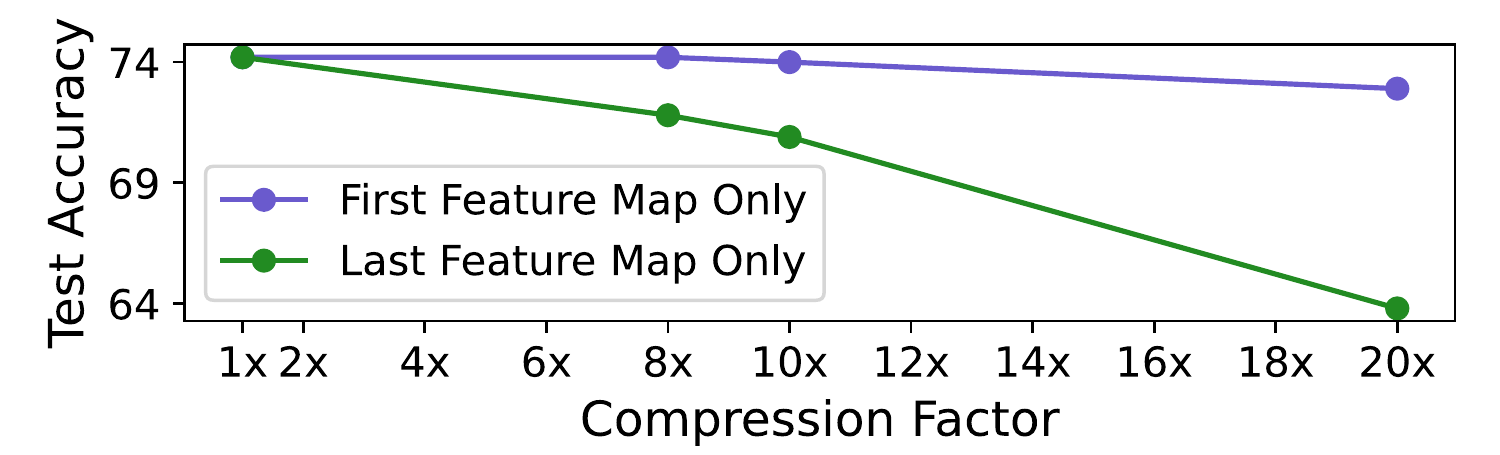}
    \caption{Compression of only the first or last feature map of VGG 19 with our Improved Projection Learning approach. The first feature map is much more compressible along the channel dimension than the last, despite the former having only 64 channels and the latter having 512 channels.}
    \label{fig:single_layer}
\end{figure}

\subsection{Early feature maps are extremely compressible}
Our first observation is that early features maps tend to be extremely compressible with our approach, typically more-so than later feature maps. Figure \ref{fig:single_layer}, for example, shows the accuracy when only the first or last feature map of VGG 19 is compressed with our method (the original network parameters frozen, and the projection layer trained for 4 epochs), for different compression factors. Accuracy drops by only $\sim 1\%$ with $20\times$ compression of the first feature map, whereas the corresponding drop is $>10\%$ for the last feature map. 

This is fortunate as earlier feature maps tend to be much larger than later ones, and so more important to compress. Furthermore, this is interesting since the natural and induced sparsity of later feature maps is generally much higher than in earlier feature maps \cite{kurtz2020inducing}, implying an increase in sparsity-based compressibilty in later layers, and thus suggesting a possible combination of our learned projections in \eqref{eq: post-activation sketch layer} and the methods tailored to sparsity such as \cite{cavigelli2019ebpc, rutishauser2020fly, kurtz2020inducing}.

\begin{figure*}[h!]
    \centering
    \includegraphics[width=\textwidth]{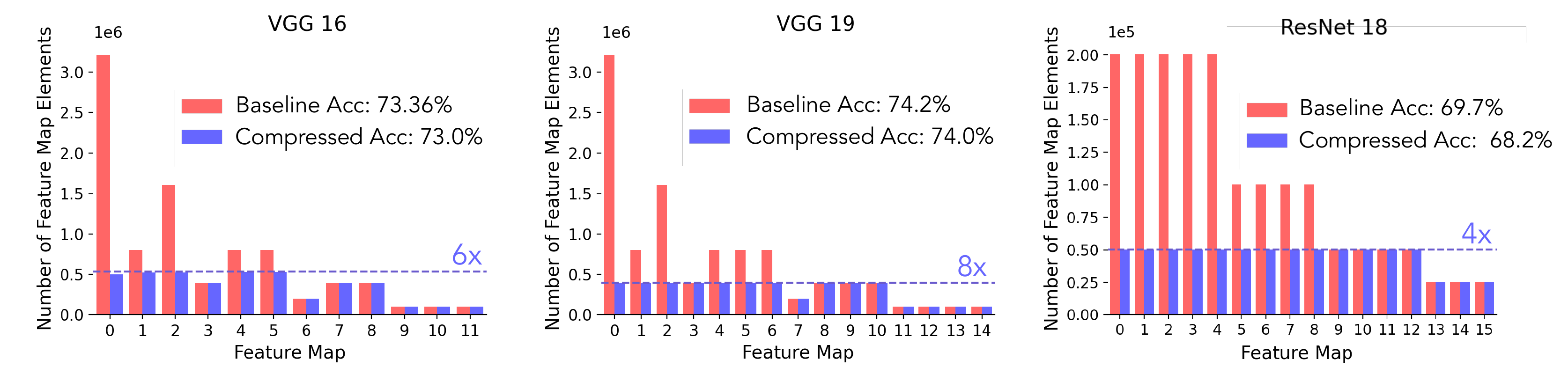}
    \caption{Results of our approach within the ceiling compression framework. The figures show the size of the feature maps in the original pre-trained networks (pink) and the compressed networks (blue). The dotted line shows the enforced ceiling, with the corresponding compression factor denoted on the its right-end. The achieved accuracy in both cases is shown in the legend.} 
    \label{fig:ceiling_compression_results}
\end{figure*}

\subsection{Ceiling compression}
We propose and study a compression paradigm we term `ceiling compression'. This is motivated by the long term goal of eliminating the need for any intermediate transfer of feature maps off-chip during the forward pass, which requires that no individual feature map should exceed a certain threshold (which in practice will be determined by available on-chip memory and the size of the model). Our target is thus to compress every feature map to at or below the level of a given ceiling, as illustrated in Figure \ref{fig:ceiling_compression_results}.

\begin{figure}[h!]
    \centering
    \includegraphics[width=0.5\textwidth]{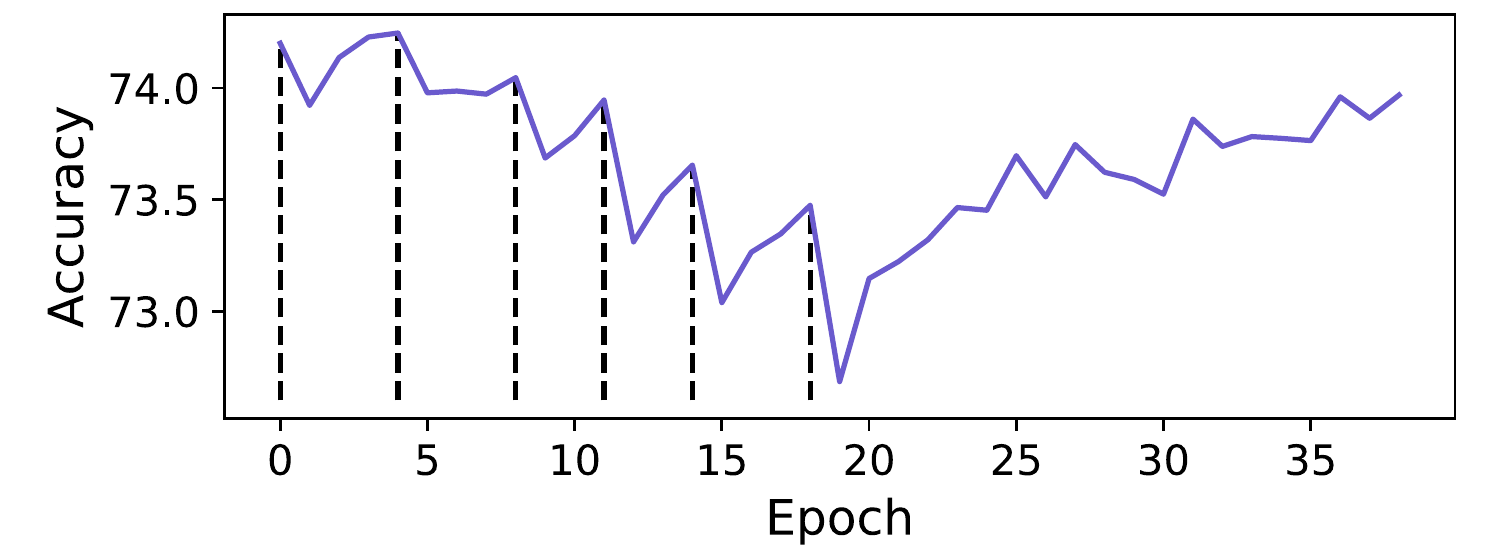}
    \caption{Combined training curve for our projected VGG 19 model. Dashed vertical lines indicate the points at which the next layer's projections were added.}
    \label{fig:training curve}
\end{figure}

\textbf{Training procedure:} The specified ceiling determines which feature maps need to be compressed and dimension of the respective projected subspaces. We hold fixed the pre-trained model, and insert the projection layers one by one, training the projection maps $S_1$ and $S_2$ for 4 epochs before adding projections to the next layer, starting from the best prior checkpoint. All prior projection layers are trainable throughout this process, and after inserting the final projection layer end-to-end training continues for 20 epochs, see Figure~\ref{fig:ceiling_compression_results}. We use SGD with 0.01 learning rate (scheduled to reduce on plateau), and do not perform hyperparameter tuning beyond this.  It is evident from Figure \ref{fig:training curve} that the training accuracy does not plateau for each projection which suggests greater accuracy is likely achievable with further training.

\textbf{Results and comparison with prior state-of-the-art:}
We applied our method to pre-trained Pytorch checkpoints of VGG~16, VGG~19, and Resnet~18 architectures. Results are shown in Figure~ \ref{fig:ceiling_compression_results}. On VGG 16 and VGG 19, $6\times$ and $8\times$ ceiling compression incurs just 0.2\% and 0.3\% drop in accuracy respectively, and on Resnet~18 a $4\times$ ceiling compression is achieved with a $1.5\%$ accuracy drop.

To the best of our knowledge, no prior work has considered something akin to the ceiling compression paradigm. Rather, prior state-of-the-art \cite{tai2022compression} targets an \textit{overall} compression rate of all feature maps -- a different goal with different constraints. In \cite{tai2022compression}, for example, projections are inserted at every layer and progressively compressed in a greedy fashion, iteratively selecting the next layer most amenable to further compression. This flexibility in how compression is distributed between layers is absent for ceiling compression. That we have not included quantisation and encoding layers limits direct comparison even further. 

Nonetheless, we report a comparison in terms of overall compression rates below for completeness. Our $6\times$ ceiling compression on VGG~16 translates to $2.1\times$ overall compression with only a  $0.36\%$ loss in accuracy.  In contrast, \cite{tai2022compression} achieve a $3.7\times$ compression rate\footnote{Note that \cite{tai2022compression} compresses \textit{pre-activation} feature maps, and the reported compression factors were presumably computed relative to the total size of all pre-activation feature maps. However, this is not the appropriate comparison, since the architectures involve a number of MaxPool layers following a Conv + ReLU, all three of which can be fused, such that, absent any feature map compression, only the output of the MaxPool ($4\times$ smaller than the preactivation feature map) need be stored.  Compression rates stated for \cite{tai2022compression} are thus recalculated to account for this.} with a loss in accuracy of $0.6\%$.  Similarly, for VGG~19 we reduce the maximum feature map size by $8\times$, which is equivalent to a $2.2\times$ overall compression and only $0.2\%$ loss in accuracy; \cite{tai2022compression} does not report results for VGG 19.  These substantial compression rates for the VGG architecture are in part due to the feature maps following the first layer being substantially larger than other layers, in contrast to ResNet, see Figure \ref{fig:ceiling_compression_results}.  For ResNet~18 a $4\times$ compression ceiling, corresponding to total compression factor $2.3\times$ results in a $1.5\%$ loss of accuracy which is notably more than the $0.4\%$ accuracy drop reported by \cite{tai2022compression} at a $2.14 \times$ total compression reduction which includes quantisation and Huffman encoding a noted above.

\section{Conclusions and further work}

Pre-trained VGG and Resnet networks, whose weights are held fixed, are shown to have their feature maps projected to prescribed memory ceilings between one fourth and one eighth the original size while losing only 0.2\% accuracy for VGG 19 and 1.5\% accuracy for Resnet 18.  This is achieved by learned projections as described in \eqref{eq: post-activation sketch layer}, which are trained by fine-tuning on the original task.  There are numerous avenues to extend this line of work such as: including quantisation and encoding techniques when storing the projected feature maps, demonstrating the efficacy of this approach on sparse networks, determining end-to-end training strategies for multiple layers at once, and implementing these methods on hardware architectures which specify the associated on-chip memory constraints.  

\bibliographystyle{IEEEbib}
\bibliography{refs}

\end{document}

%% file: math_commands.tex
\usepackage{amsmath,amsfonts,bm, amsthm}

\def\1{\bm{1}}

\DeclareMathAlphabet{\mathsfit}{\encodingdefault}{\sfdefault}{m}{sl}
\SetMathAlphabet{\mathsfit}{bold}{\encodingdefault}{\sfdefault}{bx}{n}

\newcommand{\R}{\mathbb{R}}

\DeclareMathOperator*{\argmin}{arg\,min}